\documentclass{article} % For LaTeX2e
\usepackage{iclr2026_conference,times}

\usepackage{amsmath,amsfonts,bm}

\def\eqref#1{equation~\ref{#1}}
\def\1{\bm{1}}

\def\vdelta{{\bm{\delta}}}

\def\vepsilon{{\bm{\epsilon}}}

\def\vc{{\bm{c}}}

\def\ve{{\bm{e}}}

\def\vh{{\bm{h}}}

\def\vt{{\bm{t}}}

\def\vv{{\bm{v}}}
\def\vw{{\bm{w}}}
\def\vx{{\bm{x}}}

\def\vz{{\bm{z}}}

\def\evc{{c}}

\def\mI{{\bm{I}}}

\def\mM{{\bm{M}}}

\def\mS{{\bm{S}}}

\def\mU{{\bm{U}}}
\def\mV{{\bm{V}}}

\def\mZ{{\bm{Z}}}

\def\mSigma{{\bm{\Sigma}}}
\def\mOmega{{\bm{\Omega}}}

\DeclareMathAlphabet{\mathsfit}{\encodingdefault}{\sfdefault}{m}{sl}
\SetMathAlphabet{\mathsfit}{bold}{\encodingdefault}{\sfdefault}{bx}{n}

\usepackage{microtype}
\usepackage{graphicx}
\usepackage{subcaption}
\usepackage{booktabs} % for professional tables

\usepackage[table,dvipsnames]{xcolor}
\usepackage{multirow}
\usepackage{algorithm}
\usepackage{algorithmic}

\usepackage{hyperref}
\usepackage{url}

\title{Visual Disentangled Diffusion Autoencoders: \\ Scalable Counterfactual Generation for Foundation Models}

\author{Sidney Bender \\
Machine Learning Group\\
Technische Universit\"at Berlin\\
\texttt{s.bender@tu-berlin.de} \\
\And
Marco Morik \\
Machine Learning Group\\
Technische Universit\"at Berlin\\
BIFOLD \thanks{Berlin Institute for the Foundations of Learning and Data, Berlin, Germany.} \\
\texttt{m.morik@tu-berlin.de}
}

\iclrfinalcopy % Uncomment for camera-ready version, but NOT for submission.
\begin{document}

\maketitle

\begin{abstract}
Foundation models, despite their robust zero-shot capabilities, remain vulnerable to spurious correlations and ``Clever Hans'' strategies. Existing mitigation methods often rely on unavailable group labels or computationally expensive gradient-based adversarial optimization. To address these limitations, we propose Visual Disentangled Diffusion Autoencoders (DiDAE), a novel framework integrating frozen foundation models with disentangled dictionary learning for efficient, gradient-free counterfactual generation directly for the foundation model. DiDAE first edits foundation model embeddings in interpretable disentangled directions of the disentangled dictionary and then decodes them via a diffusion autoencoder. This allows the generation of multiple diverse, disentangled counterfactuals for each factual, much faster than existing baselines, which generate single entangled counterfactuals. When paired with Counterfactual Knowledge Distillation, DiDAE-CFKD achieves state-of-the-art performance in mitigating shortcut learning, improving downstream performance on unbalanced datasets. 
\end{abstract}

\section{Introduction}
\label{section:intro}

Deep learning models, despite impressive performance on benchmarks, remain highly vulnerable to spurious correlations, often adopting ``Clever Hans'' strategies that fail to generalize out-of-distribution~\cite{lapuschkin2019unmasking, geirhos2020shortcut}. While foundation models (FMs) like CLIP~\cite{radford2021learning} have demonstrated robust few-shot capabilities, recent studies indicate they systematically encode non-causal artifacts such as background textures~\cite{kauffmann2025explainable}.

Current mitigation strategies typically rely on explicit group labels to reweight underrepresented subgroups (e.g., GroupDRO~\cite{sagawa2020distributionally}). These methods scale poorly when labels are unavailable or confounding variables are unknown. Explainable AI offers an alternative via \textit{Counterfactual Knowledge Distillation} (CFKD)~\cite{bender2023towards}, which generates counterfactuals to expose and prune reliance on confounders. However, the efficacy of CFKD is bottlenecked by the quality and speed of Visual Counterfactual Explainers (VCEs).

\begin{figure*}[t]
\centering
\includegraphics[scale=.12]{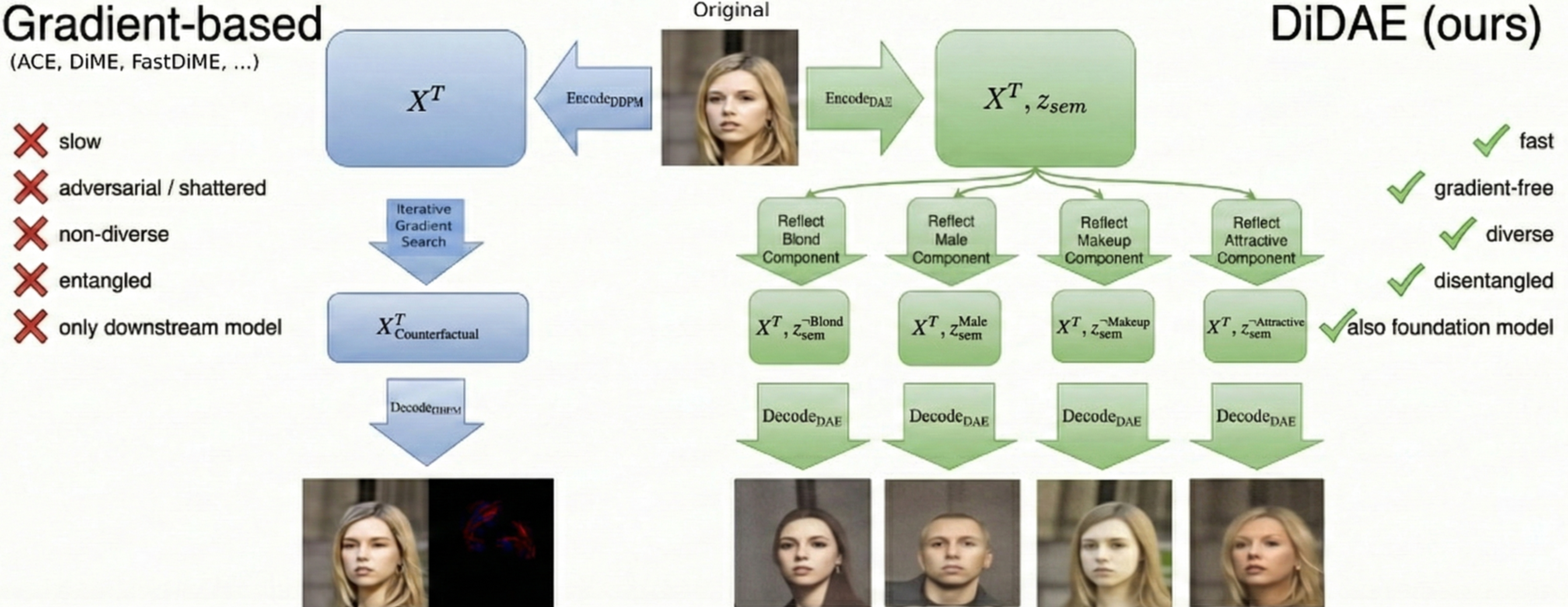}
\caption{Comparison of traditional gradient-based counterfactuals (Left) versus the proposed DiDAE approach (Right) on a CelebA classifier trained on the ``Blond Hair'' label. The label is spuriously correlated with ``Heavy Makeup'' and ``Attractive'' and anti-correlated with ``Male.'' Traditional methods require slow, iterative gradient updates through the diffusion process, often resulting in adversarial noise or entangled changes (e.g., changing hair color and eyebrows simultaneously). In contrast, DiDAE utilizes a frozen foundation model to decompose embeddings into disentangled semantic components $\vz_{sem}$. Counterfactuals are generated via simple linear reflection in this semantic space, followed by decoding via a diffusion decoder.}
\label{fig:overview}
\end{figure*}

As illustrated in Figure~\ref{fig:overview}, state-of-the-art VCEs like ACE~\cite{jeanneret2023adversarial} rely on iterative gradient-based optimization. This process is slow, often yields adversarial noise rather than semantic changes, and creates entangled edits. While recent gradient-free methods have been proposed to improve generation speed~\cite{jeanneret2024text,sobieski2024global, cao2025leapfactual}, they typically lack mechanisms for explicit semantic sparsification and diversification, limiting their utility for precise model correction.

To address these limitations, we propose \textbf{Disentangled Diffusion Autoencoders (DiDAE)}. As shown in the comparison in Figure~\ref{fig:overview} (Right), our framework wraps frozen foundation models with disentangled dictionary learning. By strictly reflecting samples along learned semantic components, DiDAE generates disentangled, diverse counterfactuals without gradient updates.

Our main contributions are as follows:

\begin{itemize}
    \item \textbf{Visual Disentangled Diffusion Autoencoders (DiDAE):} We introduce a gradient-free framework that decomposes the latent space of frozen foundation models into interpretable disentangled directions, enabling fast and precise semantic manipulation.
    \item \textbf{Scalable DiDAE-CFKD:} We demonstrate that DiDAE solves the bottleneck of Counterfactual Knowledge Distillation, enabling scalable correction of large foundation models via a pre-clustered teacher approach.
\end{itemize}

Extensive evaluations show that DiDAE significantly outperforms gradient-based baselines in generation speed and achieves superior mitigation of shortcut learning on both synthetic and natural benchmarks.

\section{Related Work}
\label{section:related}

There are 3 corpora of related work relevant for our work: the correction of models relying on spurious correlations, visual counterfactual explainers, and interpretability of foundation models.

\textbf{Spurious Correlations and Model Correction.}
Standard approaches to mitigate spurious correlations assume access to confounder annotations. Distributional robustness methods like GroupDRO~\cite{sagawa2020distributionally} and DFR~\cite{kirichenko2023last} optimize worst-group performance but struggle when minority groups are too small or unknown. XAI-based methods like P-ClArC~\cite{anders2022finding}, EGEM~\cite{linhardt2024preemptively}, and RR-ClArC~\cite{dreyer2024hope, pahde2025navigating} leverage attribution maps to penalize reliance on irrelevant features. However, as noted in recent critiques~\cite{nguyen2021effectiveness}, attributions often fail to capture geometric or global artifacts and can remain uncorrelated with the model's actual mechanics. CFKD~\cite{bender2023towards, bender2025mitigating, hackstein2025imbalanced} addresses this by using counterfactuals for data augmentation, but the generation speed and quality of the underlying counterfactuals has historically bottlenecked its performance.

\textbf{Visual Counterfactual Explainers (VCEs).}
Generating valid visual counterfactuals is a challenging inverse problem. While early methods used GANs, VAEs, or Normalizing Flows (e.g., DiVE~\cite{rodriguez2021beyond}, LatentShift~\cite{cohen2023identifying} and Diffeomorphic Counterfactuals~\cite{dombrowski2024diffeomorphic}), the field has shifted rapidly toward diffusion and flow-matching approaches. Proximal on-manifold counterfactuals can now be generated using methods such as DVCE~\cite{augustin2022diffusion}, DIME~\cite{jeanneret2022diffusion}, and Diff-ICE~\cite{pegios2025diffusion}. More specialized approaches like  CDCT~\cite{varshney2024generating} focus on generating counterfactual trajectories for concept discovery or for regression models~\cite{ha2025diffusion}. Recent advancements have also targeted semantic sparsity and computational efficiency. ACE~\cite{jeanneret2023adversarial} and FastDiME~\cite{weng2024fast} introduce mechanisms to generate semantically disentangled edits, while SCE~\cite{bender2025desiderata} explicitly optimizes for a diverse set of counterfactuals.

\textbf{Interpretability of Foundation Models.}
Recent work has focused on interpreting the latent spaces of FMs using Sparse Autoencoders (SAEs)~\cite{bricken2023towards} or decompositions like SVD. These methods decompose dense embeddings into interpretable ``concepts.'' Our work bridges this interpretability research with generative counterfactuals, using the learned disentangled dictionaries to actively generate training data for robust model correction.

\section{Methods}
\label{section:methods}
 We propose a two-stage framework: first, we construct the \textit{Visual Disentangled Diffusion Autoencoder (DiDAE)} to interpret and manipulate the latent space of a foundation model; second, we leverage this disentangled space to perform scalable model correction via \textit{CFKD}~\cite{bender2023towards} and \textit{Projection}.

%\subsection{Visual Disentangled Diffusion Autoencoder (DiDAE)}
\subsection{Disentangled Diffusion Autoencoder (DiDAE)}
\label{subsec:didae_arch}

The core of our framework is a hybrid architecture that combines a frozen discriminative encoder with a conditional generative decoder. Let $\Phi(\cdot)$ be a pre-trained foundation model (e.g., CLIP) mapping an input image $\vx \in \mathbb{R}^{H \times W \times 3}$ to a latent embedding $\vz_{\text{sem}} = \Phi(\vx) \in \mathbb{R}^{D} $.

\textbf{Latent Decomposition.}
To enable semantic manipulation, we do not operate on $\vz_{\text{sem}}$ directly. Instead, we use a disentangled invertible dictionary $\vc=\Omega(\vz_{\text{sem}})$ that decomposes the dense embedding space into coefficients $\evc_k$ aligned with canonical basis vectors $\ve_k$. Each component corresponds to a latent direction $\vv_k = \Omega^{-1}(\ve_k)$ in the embedding space. Modifying the coefficient $\evc_k$ therefore induces a change along this direction. In Section \ref{sec:latent_modification} we show two distinct ways of computing the modified embedding, while we propose two decomposition algorithms in Section~\ref{sec:disentangled_component_analysis}.

\textbf{Diffusion Decoding.}
To map these modified embeddings back to image space without the computational overhead of iterative gradient optimization as necessary for most previous VCEs \cite{augustin2022diffusion, jeanneret2022diffusion, jeanneret2023adversarial, weng2024fast, bender2025desiderata}, we employ a conditional diffusion decoder $\epsilon_\theta$. Following the Diffusion Autoencoder~\cite{preechakul2022diffusion} formulation, we encode the input $\vx$ into a semantic code $\vz_{\text{sem}}$ using the pretrained Foundation model and a stochastic spatial code $\vx_T$ using the pretrained conditional score model $\epsilon_\theta$ with DDIMInv as in~\cite{song2020denoising} (see Appendix \ref{appendix:ddim_inversion} for details). The counterfactual generation is then defined using the diffusion forward process:
\begin{equation}
    \hat{\vx} = \text{DDIM}(\vz'_{\text{sem}}, \vx_T,\epsilon_\theta)
\end{equation}
where $\vx_T$ ensures the preservation of non-semantic identity (e.g., background, texture) while $\vz'_{\text{sem}}$ alters the target attribute.

\textbf{Training.}
Crucially, we keep the foundation encoder $\Phi$ frozen to preserve its robust semantic manifold. We train only the conditional score model of the diffusion autoencoder $\epsilon_\theta$ to reconstruct $\vx$ from $(\vz_{\text{sem}}, \vx_T)$ as explained in detail in Appendix \ref{appendix:conditional_score_field}. This design allows DiDAE to inherit the few-shot capabilities of the foundation model while enabling the precise, gradient-free editing required for scalable explanation.

\subsection{Gradient-Free Analysis and Generation}
\label{sec:latent_modification}
We introduce two distinct approaches for manipulating the latent space unified in Algorithm~\ref{alg:unified_reflection}.% generates counterfactuals by strictly reflecting the sample along latent components, while Algorithm~\ref{alg:distill} generates a set of counterfactuals by projecting the sample to invert the decision boundary of a distilled classifier.

\textbf{Approach 1 (Component Reflection)} performs a rigorous semantic inversion. Instead of arbitrary amplification, it creates counterfactuals by reflecting the sample's embedding along specific component axes through the origin ($\evc_k \ve_k \to -\evc_k \ve_k$). This isolates the causal effect of reversing a specific feature (e.g., ``presence'' vs. ``absence'') while keeping the global structure intact. %Algorithm 1 would in theory also work for non-linear $\Omega$.

\textbf{Approach 2 (Distilled Boundary Inversion)} is designed to correct a specific downstream classifier $f$. First, $f$ is distilled into a linear probe $P$ within the foundation model's latent space. Then, for every component $k$ of interest, we generate a specific counterfactual by calculating an analytic projection $\vz'_{\text{sem}} \leftarrow \vz_{\text{sem}} + \vdelta$ such that the dot-product with the decision boundary is exactly inverted ($\vw^T \vz'_{\text{sem}} = - \vw^T \vz_{\text{sem}}$). When $\Omega$ is linear, this projection admits a closed-form solution $\vdelta \leftarrow \frac{-2 \vw^T \vz_{\text{sem}}}{\vw^T \vv_k} \vv_k$.

\begin{algorithm}[H]
\caption{Disentangled Diffusion Autoencoder (DiDAE)}
\label{alg:unified_reflection}
\begin{algorithmic}
\STATE {\bfseries Input:} Image $\vx$, Foundation Model $\Phi$, Diffusion Decoder $\epsilon_\theta$, 
Disentangled Dictionary $\Omega$ 
\STATE {\bfseries Optional Input :} Classifier $f$ \COMMENT{Necessary for Distilled Boundary Inversion}
\STATE {\bfseries Output:} Set of counterfactuals $\{\tilde{\vx}_k\}$
%\STATE \COMMENT{\textbf{Optional: Distilled Boundary Inversion}}
\IF{$f$ is provided} 
\STATE Fit linear probe $P(\vz)=\vw^T \vz$ such that $P(\Phi(\vx_i)) \approx f(\vx_i)$ \hfill \COMMENT{\textbf{Step 0: Distill Classifier}}
\ENDIF

%\STATE \COMMENT{\textbf{Step 1: Encode DAE}}
\STATE $z_{\text{sem}} \leftarrow \Phi(\vx)$; \quad $\vx_T \leftarrow \text{DDIMInv}(\vz_{\text{sem}},\vx,\epsilon_\theta )$ \hfill \COMMENT{\textbf{Step 1: Encode DAE}}

%\STATE \COMMENT{\textbf{Step 2.1: Calculate Components}}
\STATE $\vc \leftarrow \Omega(\vz_{\text{sem}})$  \hfill \COMMENT{\textbf{Step 2.1: Calculate Components}}

\FOR{each component $k$ of interest}
    %\STATE \COMMENT{\textbf{Step 2.2: Component-wise Reflection}}
    \STATE Let $\vv_k = \Omega^{-1}(\ve_k)$ be the direction of component $k$ in latent space
    \IF{$f$ is provided} 
        \STATE $\vdelta \leftarrow \frac{-2 \vw^T \vz_{\text{sem}}}{\vw^T \vv_k} \vv_k$ \hfill \COMMENT{\textbf{Step 2.2: Boundary inversion}}
    \ELSE 
    \STATE $\vdelta \leftarrow -2 \evc_k  \vv_k$ \hfill \COMMENT{\textbf{Step 2.2: Component reflection}}
    \ENDIF
    \STATE $\vz'_{\text{sem}} \leftarrow \vz_{\text{sem}} + \vdelta$
    \STATE $\tilde{\vx}_k \leftarrow \text{DDIM}(\vz'_{\text{sem}}, \vx_T, \epsilon_\theta)$\hfill \COMMENT{\textbf{ Step 3: Decode DAE}}
\ENDFOR
\STATE \textbf{return} $\{\tilde{\vx}_k\}$
\end{algorithmic}
\end{algorithm}

\subsection{Disentangled Component Analysis}
\label{sec:disentangled_component_analysis}
To interpret and manipulate the latent space $\vz \in \mathbb{R}^D$ of the frozen foundation models, we map dense embeddings into interpretable directions defined by a dictionary matrix $\Omega \in \mathbb{R}^{D \times D}$. Crucially, our framework is agnostic to the source of these directions: DiDAE can operate on arbitrary semantic vectors, whether derived from unsupervised decomposition, supervised alignment, or manual definition. To demonstrate this flexibility, we investigate two complementary approaches for computing $\Omega$, selected based on the availability of semantic labels.
% To interpret and manipulate the latent space of the frozen foundation models, we employ linear decomposition methods to map dense embeddings into interpretable directions. Let $\Phi(x)$ denote the frozen encoder mapping an input $x$ to a latent code $z \in \mathbb{R}^D$.

% Given a batch of embeddings $Z \in \mathbb{R}^{N \times D}$ extracted via $\Phi$, our goal is to identify a transformation matrix $\Omega \in \mathbb{R}^{D \times K}$ that disentangles the latent space into semantically meaningful axes. We investigate two complementary approaches for computing $\Omega$, depending on the availability of semantic labels.

\paragraph{Method 1: Supervised Alignment via Orthogonal Procrustes.}
When a target semantic space $\mS \in \mathbb{R}^{N \times K}$ is available (i.e., known generative factors or attribute labels), we utilize the Orthogonal Procrustes algorithm. We seek an orthogonal rotation $\mOmega = \left[ \begin{array}{c|c} \mOmega_1 & \mOmega_{\text{pad}} \end{array} \right]$ with $\mOmega_1 \in \mathbb{R}^{D \times K}$ that minimizes the element-wise difference between the rotated embeddings and the target concepts and $\mOmega_{\text{pad}} \in \mathbb{R}^{D \times (D - K)}$ to ensure full rank:
\begin{equation}
    \min_{\mOmega_{1}} \| \mZ \mOmega_{1} - \mS \|_F^2 \quad \text{subject to} \quad \mOmega^T \mOmega = \mI
\end{equation}
The optimal closed-form solution is derived via the Singular Value Decomposition (SVD) of the cross-covariance matrix $\mM = \mS^T \mZ$:
\begin{equation}
    \mM = \mU \mSigma \mV^T \quad \implies \quad \mOmega = \mV \mU^T
\end{equation}
This forces the foundation model's embeddings to align directly with user-defined concepts.

\paragraph{Method 2: Unsupervised Decomposition via SVD.}
In scenarios where ground-truth factors $\mS$ are unavailable or unknown, we employ Singular Value Decomposition (SVD) directly on the embedding matrix $\mZ$ to discover the intrinsic principal directions of variation:
\begin{equation}
    \mZ = \mU \mSigma \mV^T
\end{equation}
In this setting, we define the dictionary explicitly as the right singular vectors, $\mOmega = \mV$. While these components strictly represent directions of maximum variance rather than guaranteed semantic concepts, our framework enables their interpretation. By generating counterfactuals along specific columns of $\mOmega$, DiDAE allows us to empirically reveal the underlying visual semantics of these variations, effectively visualizing what the foundation model prioritizes in its latent representation.
%Here, the right singular vectors $V \in \mathbb{R}^{D \times D}$ serve as the transformation basis (analogous to $\Omega$). The columns of $V$ represent the directions of maximum variance in the latent space, which often correspond to disentangled features in well-structured foundation models.

\subsection{Model Correction Strategies}
\label{sec:cfkd}
Leveraging the learned disentangled dictionary, we investigate two distinct strategies for mitigating spurious correlations. 

\textbf{Projection}, which linearly removes specific component directions from the embedding space, an approach consistent with established bias mitigation literature \cite{bolukbasi2016man, chuang2023debiasing}. 
%We can fix downstream models based on the insights from looking at the counterfactuals in two different ways: With projection as related work literature already suggests doing if one found out in another way what the analyzed component does~\cite{bolukbasi2016man, chuang2023debiasing}, or with CFKD, which is only possible due to having access to counterfactuals.

\textbf{DiDAE-CFKD}, a scalable adaptation of CFKD with two options: 1) \textbf{Automatic Labeling}, where we map the foundation model's disentangled dictionary to semantic concepts once and annotate the components based on metadata, or 2) a \textbf{Preclustered Teacher}, which reduces the labeling steps for $N$ components in the foundation model, $M$ downstream models, and $K$ counterfactuals per component to just $N$ clusters of counterfactuals (compared to the $N \cdot M \cdot K$ labeling steps required by the human-in-the-loop teacher from~\cite{bender2025mitigating}).
Details can be found in Appendix \ref{appendix:cfkd}.

\section{Experiments}

In this section we describe the experimental setup starting with the used datasets over the used disentangled component analyses and the used models to the evaluation methodology.

\subsection{Datasets}
\label{sec:experimental_setup}

Following the experimental protocol established by \cite{bender2025mitigating}, we evaluate our method on two datasets: 1) the \textbf{Square Dataset:} A synthetic benchmark where the classification task involves identifying the intensity level of a small square in the foreground which can have different x- and y-positions. The spurious correlation is injected via the intensity of the background.
 2) \textbf{CelebA-Blond:} A subset of the CelebA dataset where the task is to classify the attribute ``Blond Hair''. The spurious confounder is the ``Gender'' attribute (specifically, ``Male''), which is highly correlated with the ``Non-Blond'' class in the poisoned training set, inducing a ``Clever Hans'' strategy where the model relies on gender features rather than hair color.
 
 To rigorously test model robustness against shortcut learning, we introduce a strong spurious correlation in the training data for all datasets. Specifically, we enforce a poisoning ratio of 98\%, meaning that for 98\% of the training samples, the spurious attribute is perfectly correlated with the class label (e.g., a specific background appearing with a specific class). The remaining 2\% of samples serve as counter-examples. For evaluation, we utilize a held-out test set of $N=1000$ samples for each dataset. This test set is balanced with respect to both the target class and the spurious attribute to ensure that accuracy metrics reflect true semantic learning rather than adherence to the spurious correlation. 

\subsection{Model Architectures}
\label{sec:model_architectures}

Our framework involves two distinct model components: the \textit{Foundation Model} (used as the backbone for the DiDAE) and the \textit{Student Model} (the downstream classifier being corrected).

\subsubsection{Foundation Models for DiDAE}
To enable high-fidelity, gradient-free counterfactual generation, we leverage different frozen foundation encoders $\Phi(\cdot)$ tailored to the domain of each dataset: 1) \textbf{Square (Custom):} Since this is a synthetic dataset with known generative factors, we utilize a custom foundation model trained to regress the four ground-truth latent factors: Position X, Position Y, ForegroundIntensity, and BackgroundIntensity. This ensures that the disentangled dictionary decomposition operates directly on the true disentangled manifold of the data. 2) \textbf{CelebA (CLIP):} For the natural face domain, we employ the pre-trained CLIP image encoder. Its robust zero-shot capabilities allow the DiDAE to decompose complex facial attributes into disentangled semantic directions.

\subsubsection{Student Models}
We employ two distinct types of student models to evaluate the effectiveness of our corrections: 1) \textbf{ResNet-18 (trained from scratch):} A standard CNN architecture trained on the poisoned datasets described in Section~\ref{sec:experimental_setup}. This serves as the primary subject for our CFKD experiments. 2) \textbf{Linear Probes (Foundation Model):} A linear classifier trained directly on top of the frozen foundation model's embeddings. Since the decision boundary is already linear in the foundation space, these models allow for direct application of our analytic projection methods without the intermediate distillation step required in Algorithm~\ref{alg:unified_reflection}.

\subsection{Metric Definitions}

For our evaluation we define the following 4 metrics:

\textbf{Average Group Accuracy (AGA)}
Consistent with the protocol in \cite{bender2025mitigating}, we report the Average Group Accuracy (AGA) to account for performance disparities across subgroups. The dataset is divided into disjoint groups $\mathcal{G} = \mathcal{Y} \times \mathcal{A}$ defined by the combination of the class label $y$ and the spurious attribute $a$. The AGA is calculated as the unweighted mean of the accuracy on each subgroup:
\begin{equation}
    \text{AGA}(f) = \frac{1}{|\mathcal{G}|} \sum_{g \in \mathcal{G}} \text{Accuracy}_g(f)
\end{equation}
This metric ensures that the model's performance is evaluated equally on minority groups (where spurious correlations fail) and majority groups.

\textbf{Non-Adversarial Flip Rate (NAFR)}
Following the desiderata for valid counterfactuals \cite{bender2025desiderata}, we define the Non-Adversarial Flip Rate (NAFR) as the proportion of generated counterfactuals $\tilde{x}$ that successfully flip the model prediction to the target class $y_t$ while also representing a valid semantic change according to a ground-truth oracle $O$ (i.e., the image content actually changes):
\begin{equation}
    \text{NAFR} = \frac{1}{N} \sum_{i=1}^{N} \mathbb{1}\left( f(\tilde{ \vx}_i) = y_t \land O(\tilde{\vx}_i) = y_t \right)
\end{equation}
This distinguishes robust semantic edits from adversarial attacks that flip predictions via imperceptible noise.
The oracle $O$ is just another classifier that we distill the decision strategy of our original classifier $f$ into. Because we train $O$ from scratch, this avoids the weight-specific adversarial attacks that fool $f$ also fool $O$.

\textbf{Gain}
To quantify the effectiveness of our correction strategy, we report the Gain, defined as the percentage of the performance gap closed between the baseline model and the optimal performance (100\%). Unlike simple accuracy difference, this normalized metric accounts for the varying difficulty of baselines:
\begin{equation}
    \text{Gain} = \frac{\text{AGA}(f_{\text{corrected}}) - \text{AGA}(f_{\text{baseline}})}{1 - \text{AGA}(f_{\text{baseline}})} \times 100
\end{equation}
where $f_{\text{baseline}}$ is the original model trained on poisoned data and $f_{\text{corrected}}$ is the student model after DiDAE-CFKD.

\textbf{Counterfactuals per second}
For this, we use an Nvidia-A100 with 80GB VRAM and find the batch size that is just possible when creating counterfactuals without running out of VRAM. Then we calculate counterfactuals for one batch, stop the time it takes, and divide the number of samples in the batch by the time it took. For every tuple of dataset and explainer we calculate this value for 3 batches and take the average.

\subsection{Counterfactual Evaluation Methodology}
\label{section:methodology}

We focus our evaluation on three critical metrics that assess the robustness, utility, and efficiency of the generated counterfactuals: \textbf{Non-Adversarial Flip Rate (NAFR)}, \textbf{Gain}, and \textbf{Counterfactuals per Second}.

Our evaluation pipeline proceeds in three stages. First, we utilize \textbf{Approach 1 (Reflection)} to qualitatively analyze and identify spurious components in the latent space. Second, we employ \textbf{Approach 2 (Distilled Boundary Inversion)} to generate the counterfactuals used to compute the NAFR and speed metrics. Finally, we apply the \textbf{DiDAE-CFKD} algorithm---which augments the training data with these generated counterfactuals---to measure the downstream model improvement, reported as ``Gain''.

\subsection{Downstream Improvement Evaluation Methodology}
We also evaluate how well the method can improve downstream models in terms of Average Group Accuracy in two distinct settings: 1) \textbf{ResNet-18 (Scratch):} We assess the ability of DiDAE-CFKD to correct a standard ResNet-18 architecture trained from scratch on the poisoned dataset. This evaluates the efficacy of using generated counterfactuals as data augmentation to break spurious correlations during training. 2) \textbf{Foundation Model Probing:} We evaluate the correction of linear probes trained on top of frozen foundation model embeddings (e.g., CLIP). In this setting, we test both the DiDAE-CFKD distillation and the analytic DiDAE-Proj method to measure robustness gains without fine-tuning the underlying encoder.

\section{Results}
\label{section:results_quality}

\begin{table*}[t!]
\centering \footnotesize
\caption{Quantitative comparison of our method with related work. DiDAE demonstrates superior generation speed while maintaining competitive Non-Adversarial Flip Rates (NAFR) and in terms of Gain, particularly on Foundation Models.}
\label{tab:quantiative_results}
\renewcommand{\arraystretch}{0.7}
\setlength{\tabcolsep}{4pt}
\begin{tabular}{llrccccc}
\toprule
Dataset / Model & Method & & NAFR & Gain & Counterfactuals per second \\ \midrule
\multirow{5}{*}{Square / ResNet-18}   & DiME                             & &   6.0 &   0.0 &  $\sim$ 0.02 \\
                                      & ACE                              & &   0.0 &   0.0 &  $\sim$ 0.02 \\
                                      & FastDiME                         & &   6.5 &   8.8 &  $\sim$ \underline{2.95} \\
                                      & SCE                              & &  \textbf{36.0} &  \textbf{88.8} &  $\sim$ 0.02 \\
\rowcolor{gray!10}\cellcolor{white}   & \textbf{Procrustes-DiDAE (ours)} & &  \underline{17.5} &  \underline{82.6} & $\sim$ \textbf{64.18} \\
\rowcolor{gray!10}\cellcolor{white}   & \textbf{SVD-DiDAE (ours)}        & &  \textit{17.5} &  75.7 & $\sim$ \textbf{64.18} \\ \midrule

\multirow{5}{*}{Square / Foundation}  & DiME                             & &   \underline{6.0} &   0.0 & $\sim$ 0.02 \\
                                      & ACE                              & &   0.0 &   0.0 &  $\sim$ 0.02 \\
                                      & FastDiME                         & &   5.0 &   0.0 & $\sim$ \underline{2.95} \\
\rowcolor{gray!10}\cellcolor{white}   & \textbf{Procrustes-DiDAE (ours)} & &  \textbf{22.5} &  \underline{70.4} & $\sim$ \textbf{64.18} \\
\rowcolor{gray!10}\cellcolor{white}   & \textbf{SVD-DiDAE (ours)}        & &  10.0 & \textbf{74.4} & $\sim$ \textbf{64.18} \\ \midrule

\multirow{5}{*}{CelebA-Blond / ResNet-18}       & DiME                             & &  20.0 &  18.3 &  $\sim$ 0.01 \\
                                                & ACE                              & &  26.5 &  19.9 &  $\sim$ 0.01 \\
                                                & FastDiME                         & &  12.0 &  -5.6 &  $\sim$ \underline{1.25} \\
                                                & SCE                              & &  \textbf{92.0} &  \textbf{23.4} &  $\sim$ 0.02 \\
%\rowcolor{gray!10}\cellcolor{white}             & \textbf{DinoV2-DiDAE (ours)}     & &  8.0 &  11.4 & $\sim$ \textbf{12.04} \\
\rowcolor{gray!10}\cellcolor{white}             & \textbf{Openclip-DiDAE (ours)}   & &  \underline{42.0} &  \underline{20.4} & $\sim$ \textbf{12.04} \\ \midrule

\multirow{5}{*}{CelebA-Blond / OpenClip}        & DiME                             & &  10.5 &  24.4 &  $\sim$ 0.01 \\
                                                & ACE                              & &  \underline{11.5} &  \underline{31.5} &  $\sim$ 0.01 \\
                                                & FastDiME                         & &  \underline{11.5} &  23.8 &  $\sim$ \underline{1.25} \\
\rowcolor{gray!10}\cellcolor{white}             & \textbf{DiDAE (ours)}            & &  \textbf{49.0} &  \textbf{38.5} & $\sim$ \textbf{12.04} \\

\bottomrule
\end{tabular}
\end{table*}

\begin{figure}[ht!]
    \centering
    % --- First Figure ---
    \begin{minipage}[b]{0.48\textwidth}
        \centering
        \includegraphics[width=\textwidth]{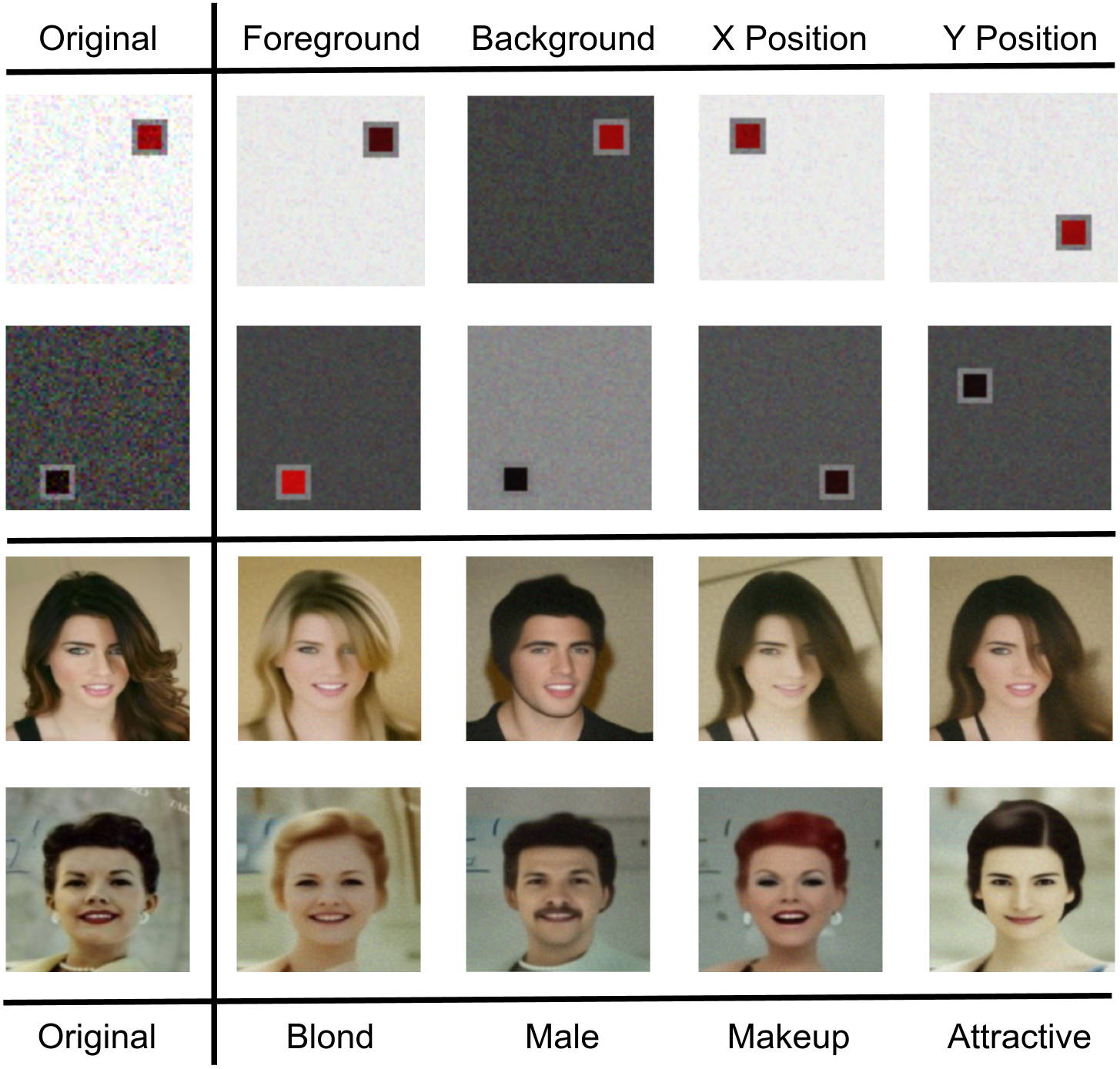}
        \caption{Visualizations of 4 example dimensions of our foundation models with DiDAE. For Square, the components correspond to the 4 latent dimensions (foreground, background, X position, Y position). For CelebA, the components reveal different attribute dimensions correlated with the ``Male'' attribute.}
        \label{fig:qualitative_results}
    \end{minipage}
    \hfill % Adds horizontal space between the figures
    % --- Second Figure ---
    \begin{minipage}[b]{0.48\textwidth}
        \centering
        \includegraphics[width=\textwidth]{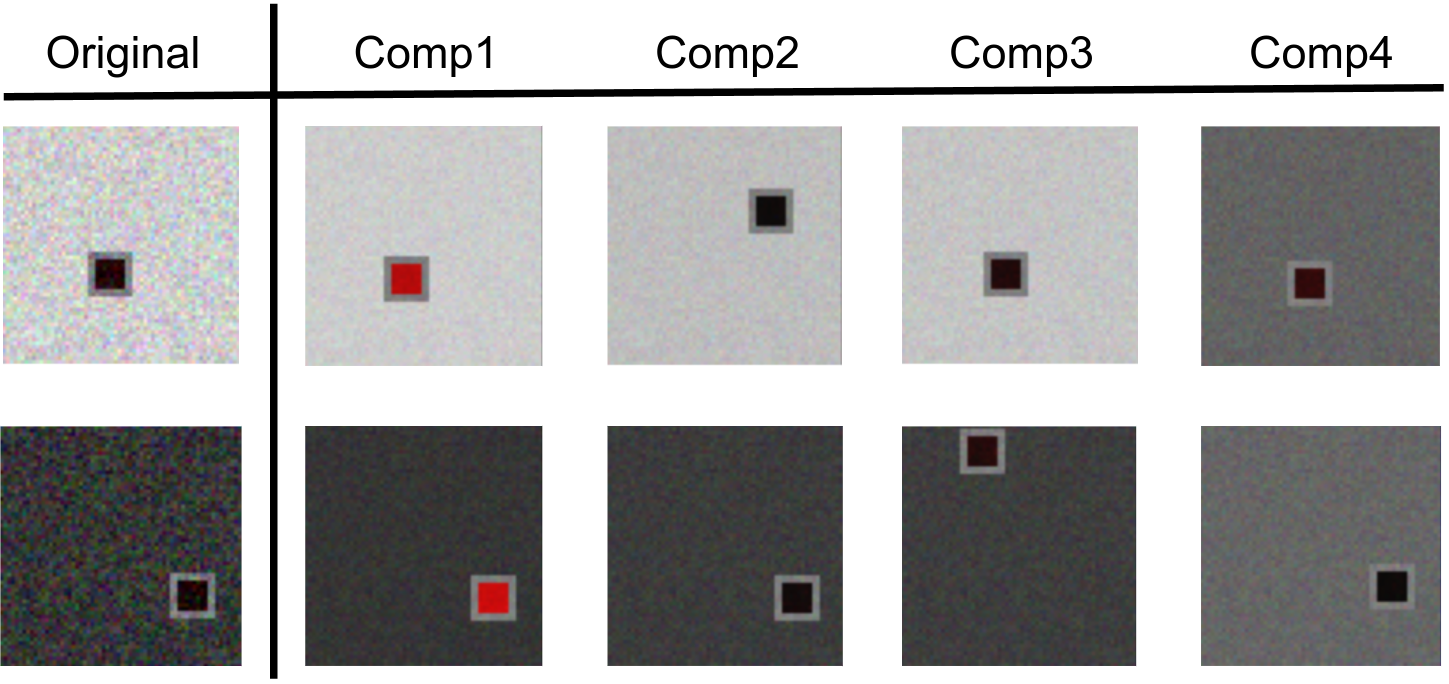}
        \caption{Visualizations of counterfactuals for the first 4 SVD dimensions found for the Square dataset in its foundation model space. One can see that Comp1 clearly corresponds to the foreground color and Comp4 clearly corresponds to the background color. Comp2 and Comp3 appear to be related to the x- and y-position, which were not disentangled perfectly. However, there is also no unique solution how the x- and y-axis could be disentangled, e.g. $\mathcal{B}_1 = \{ (0,1), (1,0) \}$ and $\mathcal{B}_2 = \{ (1,1), (-1,1) \}$ are both correct orthogonal bases.}
        \label{fig:svd_didae}
    \end{minipage}
\end{figure}
Our evaluation demonstrates that DiDAE presents a distinct trade-off relative to existing approaches: while it incurs minor reconstruction fidelity costs on standard CNNs, it offers unprecedented inference speed and unique applicability to Foundation Models.

\textbf{Computational Efficiency:} DiDAE achieves order-of-magnitude improvements in generation throughput compared to gradient-based baselines such as DiME, ACE, and SCE and is still much faster than FastDiME. As detailed in Table~\ref{tab:quantiative_results}, our method generates up to 64 counterfactuals per second, whereas optimization-based alternatives typically yield fewer than one per second.

\textbf{Quantitative Analysis (NAFR and Gain):} DiDAE significantly outperforms DiME, ACE, and FastDiME across all tasks in terms of Non-Adversarial Flip Rate (NAFR) and Gain. We attribute the limitations of the baselines to their reliance on optimization over adversarial, shattered, and non-convex loss landscapes. These characteristics impede the identification of semantic counterfactuals for entangled features with high pixel-footprints. Conversely, DiDAE's gradient-free approach facilitates clear disentanglement of latent factors. Consequently, DiDAE achieves a higher Gain by providing a sufficient volume of valid counterfactuals for Counterfactual Knowledge Distillation (CFKD), while a higher NAFR indicates that prediction flips arise from meaningful semantic edits rather than adversarial perturbations. In contrast, baselines often yield a Gain of 0 due to a failure to generate actionable counterfactuals.

\textbf{Comparison with SCE:} We observe that SCE slightly outperforms DiDAE in NAFR and Gain. We hypothesize this is due to two factors: (1) the lossy nature of DDIM inversion in DiDAE, which may result in over-smoothing and detail loss, and (2) sparse reflections occasionally exiting the variational autoencoder's support, leading to suboptimal generation. Furthermore, SCE benefits from a distillation process that smoothes classifier gradients and employs sparsification mechanisms optimized for spatially separable counterfactuals.

However, SCE exhibits critical limitations regarding scalability and applicability:
\begin{itemize}
    \item \textbf{Foundation Models:} SCE is incompatible with foundation model probing as its distillation process necessitates student randomization.
    \item \textbf{Diversity:} SCE struggles to produce more than two counterfactuals per sample, as it masks input regions that have already been modified.
    \item \textbf{Labeling Cost:} As derived in Section~\ref{sec:cfkd}, SCE-CFKD requires labeling $N \cdot M \cdot K$ counterfactuals, whereas DiDAE-CFKD requires only $N$.
    \item \textbf{Speed:} DiDAE operates up to $3200\times$ faster than SCE.
\end{itemize}

\textbf{SVD vs. Procrustes DiDAE:} While SVD-DiDAE performs marginally worse than Procrustes-DiDAE, it retains the advantage of not requiring known confounding factors. Procrustes-DiDAE, however, identifies directions that are often semantically disentangled, enabling fully automated feedback for CFKD based on metadata---a capability previously unattainable with CFKD teachers.

\textbf{Qualitative Evaluation}
DiDAE generates diverse, high-fidelity counterfactuals (see Figure~\ref{fig:qualitative_results}). The generative prior successfully decomposes dense embeddings into disentangled, interpretable ``concepts,'' allowing for the isolation of confounding signals even when spatially overlapping with causal features. This contrasts with gradient-smoothing methods (ACE, DiME, FastDiME), which fail to generate systematically diverse counterfactuals, and SCE, which is restricted to spatially non-overlapping changes. Notably, DiDAE scales generation linearly with the dimensions of the disentangled directory without quality degradation and can operate directly on foundation models without a downstream classifier.

\textbf{Performance on Downstream Model Correction Benchmarks:} Despite the reconstruction constraints of DDIM inversion on ResNet architectures, the robustness of DiDAE-generated counterfactuals translates to substantial improvements in Average Group Accuracy (AGA). As shown in Table~\ref{tab:sota_comparison}, DiDAE-CFKD leverages high-efficiency generation to aggressively augment training data. This approach yields state-of-the-art performance on the Square and CelebA benchmarks, outperforming distributional robustness methods such as GroupDRO, DFR, P-ClarC, and RR-ClarC, particularly in settings where minority groups are small or unidentified.

\textbf{Foundation Model Probing:} Table~\ref{tab:sota_comparison_foundation_model_fixing} indicates that incorporating CFKD into foundation model probing yields superior results compared to projection in the representation space. We attribute this to the redundancy of foundation models, which often encode identical information across orthogonal directions; creating a counterfactual in ambient space augments all such directions simultaneously. Furthermore, projection methods risk discarding information if the confounding direction is not perfectly disentangled from the causal feature. In contrast, augmentation remains robust to such noise, minimizing the impact of imperfect disentanglement on downstream performance.

\begin{table}[ht!]
    \centering
    \small
    
    % --- First Table ---
    \begin{minipage}[t]{0.48\textwidth}
        \centering
        \caption{Average Group Accuracy (AGA) scores comparing DiDAE-based corrections against baselines. DiDAE-CFKD consistently outperforms existing debiasing methods.}
        \label{tab:sota_comparison}
        \begin{tabular}{lcc}
            \toprule
            Method & Square & Blond \\
            \midrule
            Original & 51.1 & 74.3 \\
            GroupDRO & 61.3 & 73.0 \\
            DFR      & 52.1 & 77.5 \\
            P-ClArC  & 78.6 & 74.4 \\
            RR-ClArC & 78.6 & 74.5 \\ 
            \midrule
            \textbf{DiDAE-CFKD} & \textbf{91.9} & \textbf{78.3} \\
            \bottomrule
        \end{tabular}
    \end{minipage}
    \hfill % This adds space between the two minipages
    % --- Second Table ---
    \begin{minipage}[t]{0.48\textwidth}
        \centering
        \caption{Comparison of correction strategies on the foundation model. CFKD augmentation demonstrates superior performance compared to simple projection.}
        \label{tab:sota_comparison_foundation_model_fixing}
        \begin{tabular}{lcc}
            \toprule
            Method & Square & Blond \\
            \midrule
            Original    & 73.4 & 55.0 \\
            Projection  & 87.6 & 66.8 \\
            CFKD (ours) & \textbf{92.4} & \textbf{72.3} \\
            \bottomrule
            \addlinespace[2.5ex] % Adds invisible space to align the bottom of tables
        \end{tabular}
    \end{minipage}
\end{table}

\section{Conclusion and Future Work}
\label{section:conclusion}
In this work, we introduced Visual Disentangled Diffusion Autoencoders (DiDAE), a scalable framework that enables efficient, gradient-free counterfactual generation by wrapping foundation models with disentangled dictionary learning. Our extensive evaluation demonstrates that DiDAE outperforms gradient-based baselines by orders of magnitude in speed and non-adversarial robustness on foundation model embeddings and outperforms all explainers besides SCE on ResNets trained from scratch. By integrating DiDAE into the Counterfactual Knowledge Distillation (CFKD) loop, we achieved state-of-the-art performance in mitigating ``Clever Hans'' strategies on foundation model probes, effectively scaling robust model correction to complex, real-world benchmarks where traditional methods struggle.

Looking forward, DiDAE's gradient-free mechanism enables extensions beyond continuous image domains. By operating via semantic embedding manipulation rather than input-space optimization, this framework offers a promising path for generating counterfactuals in discrete modalities such as natural language, graphs, and protein structures. Additionally, the modular nature of our approach allows for the integration of more powerful generative backbones. Future iterations could leverage latent diffusion models like Stable Diffusion~\cite{rombach2022high} combined with advanced inversion techniques~\cite{huberman2024edit} to further enhance the realism and editability of counterfactual explanations. Finally, while our current implementation relies on linear disentanglement, extending DiDAE to non-linear representations such as sparse autoencoders \cite{gao2025scalingSAE} may enable the discovery of richer latent features, further improving the fidelity and controllability of counterfactual explanations.
%In this work, we introduced Disentangled Diffusion Autoencoders (DiDAE), a novel framework that wraps foundation models with disentangled dictionary learning to enable efficient, gradient-free counterfactual generation. We demonstrated that DiDAE satisfies key explanation desiderata—fidelity, understandability, and sufficiency—better than existing gradient-based methods, particularly in terms of non-adversarial robustness (NAFR) and image quality. Furthermore, by integrating DiDAE into the CFKD loop, we achieved state-of-the-art performance in mitigating Clever Hans strategies across synthetic and real-world benchmarks.

%For future work, the gradient-free nature of DiDAE and its reliance on semantic embedding manipulation opens the door to non-continuous domains. We anticipate that this framework can be adapted to generate counterfactuals for natural language, graphs, or protein structures, where traditional gradient-ascent attacks for counterfactual generation are often infeasible. Moreover, one can use a generative foundation model like Stable Diffusion~\cite{rombach2022high} as a backbone for the diffusion autoencoder and can use more sophisticated editing methods like DDPM inversion~\cite{huberman2024edit}.

\subsubsection*{Acknowledgments}
This work was supported by the German Ministry for Education and Research (BMBF) under Grant 01IS18037A, and by BASLEARN -- TU Berlin/BASF Joint Laboratory, co-financed by TU Berlin and BASF SE.
 
\clearpage

\bibliography{iclr2026_conference}

@inproceedings{kirichenko2023last,
  title={Last layer re-training is sufficient for robustness to spurious correlations},
  author={Kirichenko, Polina and Izmailov, Pavel and Wilson, Andrew G},
  booktitle={International Conference on Learning Representations},
  year={2023}
}

@article{lapuschkin2019unmasking,
  title={Unmasking Clever Hans predictors and assessing what machines really learn},
  author={Lapuschkin, Sebastian and W{\"a}ldchen, Stephan and Binder, Alexander and Montavon, Gr{\'e}goire and Samek, Wojciech and M{\"u}ller, Klaus-Robert},
  journal={Nature communications},
  volume={10},
  number={1},
  pages={1096},
  year={2019}
}

@inproceedings{sagawa2020distributionally,
  title={Distributionally robust neural networks},
  author={Sagawa, Shiori and Koh, Pang Wei and Hashimoto, Tatsunori B and Liang, Percy},
  booktitle={International Conference on Learning Representations},
  year={2020}
}

@article{geirhos2020shortcut,
  title={Shortcut learning in deep neural networks},
  author={Geirhos, Robert and Jacobsen, J{\"o}rn-Henrik and Michaelis, Claudio and Zemel, Richard and Brendel, Wieland and Bethge, Matthias and Wichmann, Felix A},
  journal={Nature Machine Intelligence},
  volume={2},
  number={11},
  pages={665--673},
  year={2020}
}

@inproceedings{radford2021learning,
  title={Learning transferable visual models from natural language supervision},
  author={Radford, Alec and Kim, Jong Wook and Hallacy, Chris and Ramesh, Aditya and Goh, Gabriel and Agarwal, Sandhini and Sastry, Girish and Askell, Amanda and Mishkin, Pamela and Clark, Jack and others},
  booktitle={International Conference on Machine Learning},
  pages={8748--8763},
  year={2021}
}

@inproceedings{anders2022finding,
  title={Finding and removing Clever Hans: using explanation methods to debug and improve deep models},
  author={Anders, Christopher J and Weber, Leander and Neumann, David and Samek, Wojciech and M{\"u}ller, Klaus-Robert and Lapuschkin, Sebastian},
  booktitle={Information Fusion},
  volume={77},
  pages={261--295},
  year={2022}
}

@inproceedings{dreyer2024hope,
  title={From hope to safety: Unlearning biases of deep models via gradient penalization in latent space},
  author={Dreyer, Maximilian and Pahde, Frederik and Anders, Christopher J and Samek, Wojciech and Lapuschkin, Sebastian},
  booktitle={Proceedings of the AAAI Conference on Artificial Intelligence},
  volume={38},
  pages={21046--21054},
  year={2024}
}

@inproceedings{preechakul2022diffusion,
  title={Diffusion autoencoders: Toward a meaningful and decodable representation},
  author={Preechakul, Konpat and Chatthee, Nattanat and Wizadwongsa, Suttisak and Suwajanakorn, Supasorn},
  booktitle={Proceedings of the IEEE/CVF Conference on Computer Vision and Pattern Recognition},
  pages={10619--10629},
  year={2022}
}

@article{bricken2023towards,
  title={Towards monosemanticity: Decomposing language models with dictionary learning},
  author={Bricken, Trenton and Templeton, Adly and Batson, Joshua and Chen, Brian and Jermyn, Adam and Conerly, Tom and Turner, Nick and Kundu, Cem and Denison, Carter and Hernandez, Evan and others},
  journal={Transformer Circuits Thread},
  year={2023}
}

@inproceedings{nguyen2021effectiveness,
  title={The effectiveness of feature attribution methods and its correlation with automatic evaluation scores},
  author={Nguyen, Giang and Kim, Daeyoung and Nguyen, Anh},
  booktitle={Advances in Neural Information Processing Systems},
  volume={34},
  pages={26422--26436},
  year={2021}
}

@inproceedings{rodriguez2021beyond,
  title={Beyond trivial counterfactual explanations with diverse valuable explanations},
  author={Rodriguez, Pau and Caccia, Massimo and Lacoste, Alexandre and Zamparo, Lee and Laradji, Issam and Charlin, Laurent and Vazquez, David},
  booktitle={Proceedings of the IEEE/CVF International Conference on Computer Vision},
  pages={1056--1065},
  year={2021}
}

@inproceedings{jeanneret2022diffusion,
  title={Diffusion models for counterfactual explanations},
  author={Jeanneret, Guillaume and Simon, Lo{\"\i}c and Jurie, Frederic},
  booktitle={Proceedings of the Asian Conference on Computer Vision},
  pages={858--876},
  year={2022}
}

@article{augustin2022diffusion,
  title={Diffusion visual counterfactual explanations},
  author={Augustin, Maximilian and Boreiko, Valentyn and Croce, Francesco and Hein, Matthias},
  journal={Advances in Neural Information Processing Systems},
  volume={35},
  pages={364--377},
  year={2022}
}

@inproceedings{bender2023towards,
  title={Towards Fixing Clever-Hans Predictors with Counterfactual Knowledge Distillation},
  author={Bender, Sidney and Anders, Christopher J and Chormai, Pattarawat and Marxfeld, Heike Antje and Herrmann, Jan and Montavon, Gr{\'e}goire},
  booktitle={Proceedings of the IEEE/CVF International Conference on Computer Vision},
  pages={2607--2615},
  year={2023}
}

@inproceedings{jeanneret2023adversarial,
  title={Adversarial counterfactual visual explanations},
  author={Jeanneret, Guillaume and Simon, Lo{\"\i}c and Jurie, Frederic},
  booktitle={Proceedings of the IEEE/CVF Conference on Computer Vision and Pattern Recognition},
  pages={16425--16435},
  year={2023}
}

@article{cohen2023identifying,
  title={Identifying spurious correlations using counterfactual alignment},
  author={Cohen, Joseph Paul and Blankemeier, Louis and Chaudhari, Akshay},
  journal={TMLR},
  year={2025}
}

@article{dombrowski2024diffeomorphic,
  title={Diffeomorphic counterfactuals with generative models},
  author={Dombrowski, Ann-Kathrin and Gerken, Jan E and M{\"u}ller, Klaus-Robert and Kessel, Pan},
  journal={IEEE Transactions on Pattern Recognition and Machine Intelligence},
  volume={46},
  number={5},
  pages={3257--3274},
  year={2024}
}

@inproceedings{jeanneret2024text,
  title={Text-to-image models for counterfactual explanations: a black-box approach},
  author={Jeanneret, Guillaume and Simon, Lo{\"i}c and Jurie, Fr{\'e}d{\'e}ric},
  booktitle={Proceedings of the IEEE/CVF Winter Conference on Applications of Computer Vision},
  pages={4757--4767},
  year={2024}
}

@inproceedings{varshney2024generating,
  title={Generating counterfactual trajectories with latent diffusion models for concept discovery},
  author={Varshney, Payal and Lucieri, Adriano and Balada, Christoph and Dengel, Andreas and Ahmed, Sheraz},
  booktitle={International Conference on Pattern Recognition},
  pages={138--153},
  year={2024},
  organization={Springer}
}

@inproceedings{weng2024fast,
  title={Fast diffusion-based counterfactuals for shortcut removal and generation},
  author={Weng, Nina and Pegios, Paraskevas and Petersen, Eike and Feragen, Aasa and Bigdeli, Siavash},
  booktitle={European Conference on Computer Vision},
  pages={338--357},
  year={2024},
  organization={Springer}
}

@inproceedings{ha2025diffusion,
  title={Diffusion counterfactuals for image regressors},
  author={Ha, Trung Duc and Bender, Sidney},
  booktitle={World Conference on Explainable Artificial Intelligence},
  pages={112--134},
  year={2025},
  organization={Springer}
}

@inproceedings{pegios2025diffusion,
  title={Diffusion-based iterative counterfactual explanations for fetal ultrasound image quality assessment},
  author={Pegios, Paraskevas and Lin, Manxi and Weng, Nina and Svendsen, Morten Bo S{\o}ndergaard and Bashir, Zahra and Bigdeli, Siavash and Christensen, Anders Nymark and Tolsgaard, Martin and Feragen, Aasa},
  booktitle={International Workshop on Advances in Simplifying Medical Ultrasound},
  pages={174--184},
  year={2025},
  organization={Springer}
}

@inproceedings{bender2025desiderata,
  title={Towards Desiderata-Driven Design of Visual Counterfactual Explainers},
  author={Bender, Sidney and Herrmann, Jan and M{\"u}ller, Klaus-Robert and Montavon, Gr{\'e}goire},
  booktitle={Pattern Recognition},
  year={2025}
}

@article{cao2025leapfactual,
  title={LeapFactual: Reliable Visual Counterfactual Explanation Using Conditional Flow Matching},
  author={Cao, Zhuo and Zhao, Xuan and Krieger, Lena and Scharr, Hanno and Assent, Ira},
  journal={NeurIPS},
  year={2025}
}

@article{bender2025mitigating,
  title={Mitigating Clever Hans Strategies in Image Classifiers through Generating Counterexamples},
  author={Bender, Sidney and Delzer, Ole and Herrmann, Jan and Marxfeld, Heike Antje and M{\"u}ller, Klaus-Robert and Montavon, Gr{\'e}goire},
  journal={arXiv preprint arXiv:2510.17524},
  year={2025}
}

@inproceedings{sobieski2024global,
  title={Global counterfactual directions},
  author={Sobieski, Bartlomiej and Biecek, Przemyslaw},
  booktitle={European Conference on Computer Vision},
  pages={72--90},
  year={2024},
  organization={Springer}
}

@article{hackstein2025imbalanced,
  title={Imbalanced Classification through the Lens of Spurious Correlations},
  author={Hackstein, Jakob and Bender, Sidney},
  journal={arXiv preprint arXiv:2510.27650},
  year={2025}
}

@inproceedings{huberman2024edit,
  title={An edit friendly ddpm noise space: Inversion and manipulations},
  author={Huberman-Spiegelglas, Inbar and Kulikov, Vladimir and Michaeli, Tomer},
  booktitle={Proceedings of the IEEE/CVF Conference on Computer Vision and Pattern Recognition},
  pages={12469--12478},
  year={2024}
}

@inproceedings{rombach2022high,
  title={High-resolution image synthesis with latent diffusion models},
  author={Rombach, Robin and Blattmann, Andreas and Lorenz, Dominik and Esser, Patrick and Ommer, Bj{\"o}rn},
  booktitle={Proceedings of the IEEE/CVF conference on computer vision and pattern recognition},
  pages={10684--10695},
  year={2022}
}

@article{kauffmann2025explainable,
  title={Explainable AI reveals Clever Hans effects in unsupervised learning models},
  author={Kauffmann, Jacob and Dippel, Jonas and Ruff, Lukas and Samek, Wojciech and M{\"u}ller, Klaus-Robert and Montavon, Gr{\'e}goire},
  journal={Nature Machine Intelligence 7},
  pages={412-422},
  year={2025},
  publisher={Nature Publishing Group UK London}
}

@article{linhardt2024preemptively,
  title={Preemptively pruning Clever-Hans strategies in deep neural networks},
  author={Linhardt, Lorenz and M{\"u}ller, Klaus-Robert and Montavon, Gr{\'e}goire},
  journal={Information Fusion},
  volume={103},
  pages={102094},
  year={2024},
  publisher={Elsevier}
}

@article{pahde2025navigating,
  title={Navigating neural space: Revisiting concept activation vectors to overcome directional divergence},
  author={Pahde, Frederik and Dreyer, Maximilian and Weber, Leander and Weckbecker, Moritz and Anders, Christopher J and Wiegand, Thomas and Samek, Wojciech and Lapuschkin, Sebastian},
  journal={ICLR},
  year={2025}
}

@article{song2020denoising,
  title={Denoising diffusion implicit models},
  author={Song, Jiaming and Meng, Chenlin and Ermon, Stefano},
  journal={arXiv preprint arXiv:2010.02502},
  year={2020}
}

@article{chuang2023debiasing,
  title={Debiasing vision-language models via biased prompts},
  author={Chuang, Ching-Yao and Jampani, Varun and Li, Yuanzhen and Torralba, Antonio and Jegelka, Stefanie},
  journal={arXiv preprint arXiv:2302.00070},
  year={2023}
}

@article{bolukbasi2016man,
  title={Man is to computer programmer as woman is to homemaker? debiasing word embeddings},
  author={Bolukbasi, Tolga and Chang, Kai-Wei and Zou, James Y and Saligrama, Venkatesh and Kalai, Adam T},
  journal={Advances in neural information processing systems},
  volume={29},
  year={2016}
}

@article{gao2025scalingSAE,
  author       = {Leo Gao and
                  Tom Dupr{\'{e}} la Tour and
                  Henk Tillman and
                  Gabriel Goh and
                  Rajan Troll and
                  Alec Radford and
                  Ilya Sutskever and
                  Jan Leike and
                  Jeffrey Wu},
  title        = {Scaling and evaluating sparse autoencoders},
  journal    = {{ICLR}},
  year         = {2025},
}
\bibliographystyle{iclr2026_conference}

\appendix

\section{Counterfactual Knowledge Distillation (CFKD)}
\label{appendix:cfkd}

To mitigate the reliance on spurious correlations, we employ Counterfactual Knowledge Distillation (CFKD) \cite{bender2023towards}, adapting it to leverage the high-throughput generation capabilities of our proposed DiDAE framework. CFKD is a data augmentation strategy that distills true causal mechanisms from a teacher into a student classifier $f$ by exposing it to semantically manipulated counterfactuals. 

As detailed in Algorithm~\ref{alg:cfkd}, the CFKD process assumes four primary components: (i) a trained student classifier $f$, (ii) a visual counterfactual explainer (in our case, DiDAE), (iii) a training dataset $\mathcal{D}$, and (iv) a teacher $t$ (which can be a human-in-the-loop, an oracle, or our scalable pre-clustered approach). During each iteration, the VCE generates a counterfactual $\tilde{x}$ targeted at a specific class $y_{\text{target}}$. The teacher then evaluates whether the generation process successfully altered the underlying causal feature (a ``True'' counterfactual) or merely altered non-causal/spurious artifacts (a ``False'' counterfactual). If the causal feature remains unchanged (False), the generated image $\tilde{\vx}$ is injected back into the training dataset with its original factual label $y$, thereby teaching the model to ignore the spurious transformations. Otherwise it is discarded. The classifier $f$ is subsequently fine-tuned on this augmented dataset. Because DiDAE produces highly interpretable, disentangled image-counterfactual pairs at scale, this feedback loop can efficiently correct the student model's decision boundaries without the traditional computational bottlenecks.

\begin{algorithm}[H]
\caption{Counterfactual Knowledge Distillation (CFKD)}
\label{alg:cfkd}
\begin{algorithmic}[1]
\STATE {\bfseries Input:} Trained student classifier $f$, training dataset $\mathcal{D}$, Visual Counterfactual Explainer (DiDAE), teacher $t$, number of iterations $n$
\STATE {\bfseries Output:} Fine-tuned classifier $f'$
\STATE $\mathcal{D}_{\text{aug}} \leftarrow \emptyset$
\FOR{each $(\vx, y) \in \mathcal{D}$}
    \STATE Select target label $y_{\text{target}} \neq y$
    \STATE Generate counterfactual image $\tilde{\vx}$ using DiDAE based on $\vx$ and $y_{\text{target}}$
    \STATE $eval \leftarrow t(\vx, \tilde{\vx})$ \hfill \COMMENT{Teacher decides if $\tilde{\vx}$ is a True or False counterfactual}
    \IF{$eval$ is False}
        \STATE $\mathcal{D}_{\text{aug}} \leftarrow \mathcal{D}_{\text{aug}} \cup \{(\tilde{\vx}, y)\}$ \hfill \COMMENT{Retain original label}
    \ENDIF
\ENDFOR
\STATE Retrain the classifier $f$ on $\mathcal{D} \cup \mathcal{D}_{\text{aug}}$
\STATE \textbf{return} $f$
\end{algorithmic}
\end{algorithm}

\section{Training the Conditional Score Field}
\label{appendix:conditional_score_field}

As introduced in Section~\ref{subsec:didae_arch}, our Disentangled Diffusion Autoencoder (DiDAE) utilizes a conditional generative model to map manipulated semantic embeddings back to the image space. While standard diffusion autoencoders optimize both the semantic encoder and the decoder jointly, our framework preserves the robust semantic manifold of the foundation model by keeping the encoder $\Phi(\cdot)$ strictly frozen. Consequently, we only train the parameters $\theta$ of the conditional score field, which we parameterize via a noise prediction network $\epsilon_\theta$.

The conditional score field models the reverse generative process to match the inference distribution $q(\vx_{t-1} |\vx_t, \vx_0)$ conditioned on the semantic embedding $\vz_{\text{sem}} = \Phi(\vx_0)$:
\begin{equation}
    p_\theta(\vx_{0:T} \mid \vz_{\text{sem}}) = p(\vx_T) \prod_{t=1}^T p_\theta(\vx_{t-1} \mid \vx_t, \vz_{\text{sem}})
\end{equation}

To train the conditional score field, we optimize the simplified variational bound objective \cite{preechakul2022diffusion}. Unlike standard architectures where gradients flow back through the encoder parameters $\phi$, our training objective is optimized solely with respect to the score field parameters $\theta$:
\begin{equation}
    L_{\text{simple}} = \sum_{t=1}^T \mathbb{E}_{\vx_0, \vepsilon_t} \left[ \| \epsilon_\theta(\vx_t, t, \vz_{\text{sem}}) - \vepsilon_t \|_2^2 \right]
\end{equation}
where $\vepsilon_t \sim \mathcal{N}(0, I)$, the noisy image is defined as $\vx_t = \sqrt{\alpha_t} \vx_0 + \sqrt{1 - \alpha_t} \vepsilon_t$, and $T$ is the total number of diffusion steps (e.g., $1000$). Note that the stochastic subcode $\vx_T$ is not required during the training phase.

Following the architecture detailed in \cite{preechakul2022diffusion}, the conditional score field $\epsilon_\theta$ is implemented as a modified U-Net. We inject both the timestep $t$ and the frozen semantic conditioning $\vz_{\text{sem}}$ into the network using adaptive group normalization (AdaGN) layers. These layers extend standard group normalization by applying channel-wise scaling and shifting to the normalized feature maps $\vh \in \mathbb{R}^{c \times h \times w}$:
\begin{equation}
    \text{AdaGN}(\vh, t, \vz_{\text{sem}}) = \vz_s (\vt_s \text{GroupNorm}(\vh) + \vt_b)
\end{equation}
where $\vz_s \in \mathbb{R}^c = \text{Affine}(z_{\text{sem}})$, and $(\vt_s, \vt_b) \in \mathbb{R}^{2 \times c} = \text{MLP}(\psi(t))$ is the output of a multilayer perceptron processing the sinusoidal encoding $\psi(t)$ of the current timestep. 

By freezing $\vz_{\text{sem}}$ and utilizing AdaGN, the conditional score field learns to generate high-fidelity image variations perfectly guided by the semantic boundaries defined entirely by the pre-trained foundation model.

\section{DDIM Inversion and Counterfactual Decoding}
\label{appendix:ddim_inversion}

To isolate the semantic information from the low-level spatial and texture details, we rely on the deterministic generation process of the Denoising Diffusion Implicit Models (DDIM) \cite{song2020denoising}. As demonstrated by Song et al., the DDIM sampling process can be formulated as an Euler integration for solving ordinary differential equations (ODEs). This ODE perspective allows us to run the generative process in reverse, mapping an input image $\vx_0$ to a stochastic spatial code $\vx_T$ without introducing random noise.

In our conditional architecture, the DDIM inversion phase (referred to as $\text{DDIMInv}$ in Algorithm~\ref{alg:unified_reflection}) utilizes the conditional score field $\epsilon_\theta$ guided by the original, frozen semantic embedding $\vz_{\text{sem}} = \Phi(\vx_0)$. We discretize the ODE and reverse the time steps from $t=0$ to $T$. For a given step from $t$ to $t+\Delta t$, the inversion update is computed as:
\begin{equation}
    \frac{\vx_{t+\Delta t}}{\sqrt{\alpha_{t+\Delta t}}} = \frac{\vx_t}{\sqrt{\alpha_t}} + \left( \sqrt{\frac{1 - \alpha_{t+\Delta t}}{\alpha_{t+\Delta t}}} - \sqrt{\frac{1 - \alpha_t}{\alpha_t}} \right) \epsilon_\theta(\vx_t, t, \vz_{\text{sem}})
\end{equation}
This deterministic trajectory encodes the observation $\vx_0$ into the latent spatial representation $\vx_T$. Because $\vz_{\text{sem}}$ captures the high-level semantic attributes, $\vx_T$ effectively encodes the residual structural and identity-preserving information (e.g., background and pose) that is orthogonal to the foundation model's embedding.

Once the semantic embedding is manipulated into a counterfactual direction $\vz'_{\text{sem}}$ (e.g., via component reflection or boundary inversion), we generate the final counterfactual image $\tilde{\vx}$ by solving the ODE forward in time (from $T$ back to $0$). This decoding phase uses the exact same deterministic integration method, but is now conditioned on the edited semantic embedding $\vz'_{\text{sem}}$:
\begin{equation}
    \frac{\vx_{t-\Delta t}}{\sqrt{\alpha_{t-\Delta t}}} = \frac{\vx_t}{\sqrt{\alpha_t}} + \left( \sqrt{\frac{1 - \alpha_{t-\Delta t}}{\alpha_{t-\Delta t}}} - \sqrt{\frac{1 - \alpha_t}{\alpha_t}} \right) \epsilon_\theta(\vx_t, t, \vz'_{\text{sem}})
\end{equation}

By utilizing the inverted noise $\vx_T$ as the starting point and conditioning the generative path on $\vz'_{\text{sem}}$, the newly decoded image $\tilde{\vx}$ smoothly adopts the edited semantic concepts while rigorously preserving the non-semantic identity of the original image $\vx_0$.

\section{Hyperparameters}

The hyperparameters for training the conditional score field and for creating the counterfactuals with the help of DDIM inversion follow~\cite{ha2025diffusion}. The implementation follows it as well with the change, that we injected and froze the semantic encoder. For running CFKD we used the same hyperparameters as in~\cite{bender2025mitigating} which already had configurations for the ResNet18 runs. For the runs based on foundation models we just swapped out the model and kept everything else the same. We also did not do any hyperparametersearch for CFKD on top of this and ran the experiments for all explainers with the same parameters.

\end{document}